# Exploring Personality and Online Social Engagement: An Investigation of MBTI Users on Twitter

Partha Kadambi

## Abstract

Text-based personality prediction by computational models is an emerging field with the potential to significantly improve on key weaknesses of survey-based personality assessment. We investigate 3848 profiles from Twitter with self-labeled Myers-Briggs personality traits (MBTI) – a framework closely related to the Five Factor Model of personality – to better understand how text-based digital traces from social engagement online can be used to predict user personality traits. We leverage BERT, a state-of-the-art NLP architecture based on deep-learning, to analyze various sources of text that hold most predictive power for our task. We find that biographies, statuses, and liked tweets contain significant predictive power for all dimensions of the MBTI system. We discuss our findings and their implications for the validity of the MBTI and the lexical hypothesis, a foundational theory underlying the Five Factor Model that links language use and behavior. Our results hold optimistic implications for personality psychologists, computational linguists, and other social scientists aiming to predict personality from observational text data and explore the links between language and core behavioral traits.

## 1. Introduction

The modeling of personality traits offers a powerful approach to understanding the diversity of human behavior at the level of the individual and in theorizing the emergence of complex society-wide phenomena. A complete understanding of several social phenomena – from the formation of specific political beliefs to the diversity of lifestyle and information preferences – lies downstream of an understanding of the nature of core cognitive/psychological preferences of persons. Personality psychometrics attempts to discover these crucial axes of human behavior and devise methods to measure these dimensions.

A growing number of political analysts, social psychologists, and behavioral economists are incorporating models of personality in their studies to account for the effects of individual differences in the decision making attitudes of people. These differences are important in determining social outcomes[1]. Particular measures of personality differences such as the five-factor model (FFM) have demonstrated significant correlations with several indicators of achievement and social outcome, including college GPA and job satisfaction[2, 3]. The Myers Briggs Type Indicator (MBTI)[4] – which comprises of four scales that all directly correlate with FFM traits[5] – is closely related and enjoys high popularity in public culture and has shown to correspond to qualitative assessments of communication styles and other behavior[6, 7]. Personality psychometrics can serve as a useful tool in understanding the links between specific social outcomes germane to social scientists and individual behavior. They also provide highly efficient methods to control for variance in individual behavior in social science experiments that involve human subjects.

Psychometrics has conventionally employed survey-based personality questionnaires in its methodology. The increasing availability of rich social data online and the development of computational syntactic parsers capable of analyzing text at scale has spurred interest in computationally enhanced methods for text-based personality psychometrics. A computational approach can leverage a variety of statistical and machine-learning methods to extract signals of personality traits from text data; these include methods for exploratory analysis wherein we can measure semantic associations between personality labels and social data and supervised methods to predict personality trait labels or scores from the data.

A computationally-enabled approach has certain key advantages over a questionnaire-based design. Firstly, a text-based observational approach does not 'disturb' the subject – the subject's behavior is closer to their natural state in this case as compared to when prompted by a survey. Secondly, text data is a window into the social behavior of a person – it tends to capture behavioral with more granularity than typical personality instruments that use Likert scales (which have largely focused on understanding relationships between macro-outcomes such as education/career success and personality traits). Thirdly, it is vastly cheaper and more scalable than running similarly powered human subject experiments.

Advancements in Natural Language Processing in the past few years have made it possible for computers to extract semantic information from text to a degree not far behind the capabilities of human beings. While text-based psychometrics has been demonstrated as a possibility for longer than a decade, the recent wave of progress in NLP can likely empower much better text-based predictive models of personality. This opens the door to a new class of questions that probe the link between semantic patterns underlying text-based social communication and personality traits.

While these advancements can be framed as 'new', the principles at the core of this approach are strongly cemented in the foundations of some of the most influential frameworks of personality such as the FFM. The FFM was originally conceived from the lexical hypothesis[8] – the idea that that language or lexical items can be grouped into distinct psychological groups that correspond to separable dimensions of human behavior. We advance the view that computationally leveraged large-scale text analyses can provide crucial validation of some of the fundamental theory underlying dominant models of personality. In our view, text-based approaches are not principally sub-standard to survey-based approaches to personality psychometrics (which are often termed as the gold standard for ground truth); rather, we believe that text-based approaches directly investigate empirical claims underlying the construction of personality traits construed from the lexical hypothesis. This precise position is adopted by Kulkarni et al. (2018)[9].

This paper aims to understand how personality traits manifest in social communications online through an analysis of the engagement and information consumption patterns of Twitter users. We investigate social communications of Twitter users who self-label their personality type within the MBTI personality framework. This includes exploring the semantic spaces occupied by the social discourse *generated* by different personality types and also how personality traits relate to information *consumption* preferences online.

Our second objective is to validate key constructions and theory underlying the MBTI system by analyzing the prediction accuracies of deep-learning based models tasked with predicting user MBTI traits (given other user data). These models, which are capable of extracting contextually aware semantic information, can shed light on specific user behaviors that carry signals for personality traits. The ability of these models to label user traits correctly is indicative, to some extent, of the validity of the trait in question itself.

## 2. Background and related literature

Text-based approaches are relatively new to personality psychometrics, as methods capable of analyzing text data to significant depth have only been developed recently. Quercia et al. (2011)[10] were amongst the early pioneers of using Twitter data to predict personality. That personality traits are strongly linked to social media interaction patterns was best demonstrated by Kosinski et al. (2013)[11] through large-scale analysis of Facebook data. The same year, Schwartz et al. (2013)[12] showed that an open-vocabulary approach (as opposed to looking for pre-established linguistic markers) could succeed in capturing behavioral and demographic attributes. In 2015, Park et al.[13] further explored the possibility of assessing personality through language expressed on social media and found significant correlations between the Big 5 traits and language use online.

While these studies have validated the language-based social media approach to personality assessment, the prediction of personality traits using machine learning and Natural Language Processing is still an active and relatively unexplored research area. This paper aims to improve prior models by adopting sophisticated NLP methods that have been developed in the last two years and using the largest dataset of its kind for the Twitter platform. This would serve the purpose of validating a text-based approach to predict MBTI traits and also shed light on the validity of the MBTI framework itself. We also aim to understand how multiple aspects of social engagement – self-descriptions, information consumption (the kind of information consumed by the user) and information generation (original messages shared by the user) – are related to personality traits. This would investigate how various *kinds* of text-based digital trace, particularly on Twitter, relate to personality traits - a question relatively untouched by prior research.

## 2.1 MBTI and text-based predictive approaches

The MBTI is a commercially and publicly popular psychometric framework that classifies persons into one of 16 'types'. It measures four traits (Introversion/Extraversion, Sensing/Intuition, Thinking/Feeling, and Judging/Perceiving) and assigns a letter label to each trait dichotomy to produce a 4 letter personality 'type'. The four dimensions are briefly described here:

• Extraversion/Introversion – whether you direct energy outwardly or inwardly, whether people are energized by spending time with other people or by alone time (similar to FFM extraversion)

• Intuition/Sensing – the kind of information people pay attention to; whether you prefer an abstract or concrete worldview (similar to FFM openness)

• Thinking/Feeling - preference for decision making strategy; prioritizing people-oriented or objectively oriented decision-making (correlates with FFM agreeableness)

• Judging /Perceiving – whether you prefer clear goal-oriented scheduling or are more flexible and prospecting for new experiences (correlates with FFM conscientiousness)

While this rigid typology has attracted criticism within academic circles[14] it should be noted that it is a close cousin of the FFM – all four dimensions correlate significantly with four FFM scales[5]. It has been pointed out that the FFM is a superior alternative to the MBTI – however, this is only partly true. While the FFM is a generally better accepted and well-validated tool, it is important to note that the underlying theory behind the MBTI and the FFM are very different. While these frameworks are ostensibly comparable, the interpretational structure underlying the MBTI is far more complex than that underlying the FFM – extensive qualitative analysis and anecdotal observations are built into the theory underlying the MBTI. MBTI is based on investigative work by Carl Jung[15] while the FFM is based on the lexical hypothesis and derived from factor analysis. The MBTI simplifies the theory of psychological functions advanced by Carl Jung by framing it as a typology - by doing so, it distorts the concepts it was originally built on (for example, Carl Jung's psychological functions are not dichotomous to the degree espoused by the MBTI, nor are they directly indicative of individual letters in the 4-letter MBTI code). It is interesting to note that the FFM and the MBTI are highly similar – given that innumerable possibilities of personality frameworks – despite being developed by two entirely different paradigms. This should give us more confidence that some common behavioral patterns are indeed being captured by these frameworks.

There are two primary justifications for the choice of using MBTI (over the FFM) as the personality framework for purposes of understanding the link between personality traits and social communications online:

1) **Availability of data**: MBTI personality data of users on online social networks is far more accessible than FFM personality data. The MBTI is a hugely popular personality framework and large communities of persons who self-identify their MBTI can be found on most large social media websites – including Twitter, Reddit, Facebook, and YouTube. No such communities exist for any of the models that fall under the umbrella of FFM (OCEAN, Big 5, etc.). This makes the FFM unsuitable for large-scale social media-based projects as questionnaire-based approaches will likely be expensive and possibly require months of effort to collect similar scales of data. On the other hand, APIs can be leveraged to both search for users who self-label their personality and also collect their social engagement information if publicly accessible; data collection would be far cheaper both in terms of time and money. This paper analyzes about 2M tweets corresponding to about 3800 self-labeled Twitter profiles; running a personality survey (required for measuring FFM dimensions) for these many profiles would be a costly endeavor.

2) **Validating the MBTI (and the lexical hypothesis):** While the FFM has typically been shown to demonstrate better test-retest validity than the MBTI instrument, this may not be a reflection of the validity of the theory underlying MBTI; rather, this may be because the MBTI instrument is itself unfaithful to its underlying theory or simply a significantly error-prone approach to analyzing personality. Critical comparisons between the lexical hypothesis and Carl Jung's theory of psychological functions are few and far between in existing literature (not to suggest that these frameworks are necessarily incompatible). The theoretical differences between these models create problems for direct comparisons between these models. In sum, the MBTI may be a poorer *instrument* than the FFM, but it is rather an apples to oranges comparison. There is some early empirical evidence – from neuroscience – suggesting neural correlates to Carl Jung's theories of type[16]. There is sufficient principled cause to investigate the validity of the theory underlying MBTI itself. Also, note that validating the presence of MBTI traits would be consistent with the finding that the MBTI attitudes correlate with FFM dimensions; moreover, as this is a purely lexical approach, any predictive signals of traits would also speak to the validity of the lexical hypothesis itself.

## 2.2 Review of related literature

Text-based prediction of MBTI labels is currently an active research area and seems to attract more research interest than text-based FFM trait prediction, primarily due to the greater availability of MBTI data. In 2015, Plank and Hovy[17] analyzed 1.2M tweets annotated by MBTI type and found that only the Introvert/Extrovert and Thinking/Feeling traits were predictable with a chance above random. However, they used features and techniques that, by today's standards, are fairly basic – they used a logistic regression classifier on bigrams, gender, and meta-features such as tweet counts etc. In 2016, Verhoeven et al.[18] used word and character n-grams from labeled Twitter posts from users of six languages to predict MBTI trait labels. Only a small fraction of the trait classifiers (across six languages) they test perform better than the majority baseline. They report that the classifiers that do perform better than random are for the Introvert/Extravert and Thinking/Feeling dimensions; this aligns with the findings of Plank and Hovy.

With the introduction of BERT (Bidirectional Encoder Representations from Transformers) – a deep-learning-based contextual embedding model – by Devlin et al. in late 2018[19], research incorporating BERT to predict MBTI traits started to emerge. Mehta et al. (2020)[20] used an ensemble model composed of BERT and a multi-layer perceptron to predict MBTI labels from a Kaggle dataset composed of posts by users from the personality forum *personalitycafe.com* with self-identified MBTI labels, reporting significant prediction performance for the Thinking/Feeling and Perceiving/Judging dimensions and also SOTA results for the open dataset they used. A similar dataset was analyzed by Keh et al. (2019)[21] using BERT; they found that BERT performed better at I/E and T/F dimensions.

Results from these studies, taken as a whole, suggest that while MBTI personality prediction is not an easy task for NLP-based ML models, some MBTI traits display linguistic signal. However, results based on *personalitycafe.com* posts must be interpreted with scepticism. Firstly, as the forum is dedicated to discussing personality, posts by users are often heavily laden with explicit and implicit references to their (own) personality type, leading to a form of data leakage. Secondly, the discourse captured by these posts reflects a highly specific form of contextual engagement that few persons would ever take part in. As this form of engagement itself is not generalizable to general social media engagement, the scope of models built on such data is highly limited. These are serious issues that, in our view, are grounds for significant scepticism of claims that performance on such datasets reflects the ability of ML models to capture personality traits from text or validates the system of traits in question. Even researchers working on Twitter data express pessimism: in a review of automatic text-based methods to predict MBTI personality, Štanjer and Yenikent (2020)[22] conclude with a rather grim note – they argue that '*Twitter data simply do not make for a good dataset for the MBTI personality detection, and even more, that purely textual data do not exhibit clear linguistic (either content-based or stylistic) signals for it.*'

As we will see in the results section, this view is at least partially incorrect; the authors have likely framed their argument in the context of some Twitter studies that fail to capture important MBTI dimensions and generally suffer from poor prediction accuracy. This paper will argue that improvements in the quality of the (Twitter) data used and the prediction techniques obviate concerns that Twitter data is fundamentally incomplete or inappropriate for predicting MBTI traits. We do this by building a new Twitter dataset of users who self-label their MBTI type and using powerful text-embedding techniques that are capable of contextually aware syntactic parsing to predict type.

## 2.3 Research objectives

The research objectives of this paper are:

1) To build predictive models of MBTI traits that improve on prior research results and understand its implications for the validity of the MBTI schema as well as that of the lexical hypothesis.
2) To perform an explorative analysis of user tweets and liked tweets to qualitatively understand how the diversity of social media behaviour relates to personality traits.

## 3. Methods and discussion

In brief, we mine the Twitter data of profiles who identify their MBTI type and train a deep-learning-based NLP model (BERT) to classify MBTI traits and personality types. We also perform exploratory analysis involving embedding-based dimensionality reduction and visualization using UMAP. We apply these techniques across a comprehensive array of data – user tweets (statuses), user liked tweets, and user biographies – to understand how user discourse and information preferences align with MBTI personality traits.

### 3.1 Data and discussion

We mine Twitter data of 3848 persons. This data includes user tweets, tweets liked by the user, and user biography. Users with an abnormally high number (>1000) of follows were excluded on suspicion that they are bots. All text was cleaned of MBTI labels to ensure no cross-contamination of training data and labels. Mentions (tokens starting with '@') and URLs were removed. Text data is cleaned through tokenization/lemmatization or other NLP techniques depending on the requirements of the analytical methods used downstream.

Note that the MBTI labels are user-provided; hence, these labels are prone to erroneous identification. We expect that most users will not have arrived at these labels by undertaking a well-validated test administered by professionals. We also expect some users to have declared personality traits that they *aspire* to have rather than reflect true behavior. While both these concerns are valid, we think the issue of label accuracy is partially mitigated as these users are confident of their identities in so far as they are invested in the MBTI theory and feel confident to share their type publicly. Even if there are errors in self-identification, we expect some signal related to true MBTI type (if it exists) to remain. In the worst case, we expect that the results of this study generalize only to *self-identified* personality. Finally, note that due to the presence of these issues, the prediction accuracies reported by our analysis are a *lower bound* of the true prediction accuracies (that is if the labels were generated by a perfect instrument). Hence, any significant predictive performance generated by our models has optimistic implications for this line of research.

Our data also suffers from some degree of self-selection; this is immediately observable in the distribution of Intuitive and Sensing types in the data (see table 1) – Intuitive types are disproportionately dominant in our dataset based on a national population sample[23]. This is a problem as it leads to unbalanced datasets, but we correct this through sampling. This finding is a signal that *some* validity exists in the labels, as this is in line with the theory underlying intuitive/sensing preferences. Prior studies on other social media websites have also encountered precisely this pattern.

We split the data into train/test sets on a 90/10 basis. To run experiments, we further sample the data to ensure balanced classes of labels (note that this creates unequal N across tasks).

| Personality Trait | N |
|---|---|
| Introverts (I) | **1985** |
| Extraverts (E) | **1863** |
| Intuitive (N) | **2279** |
| Sensing (S) | **1569** |
| Feeling (F) | **1986** |
| Thinking (T) | **1862** |
| Prospecting (P) | **1713** |
| Judging (J) | **2135** |

Table 1: No. of profiles

## 3.2 BERT

Developed by researchers from Google in late 2018, BERT represents the latest paradigm in deep-learning approaches to Natural Language Processing. Composed of 12 layers of Transformer language encoder-decoders and trained on a bidirectional language modeling task, BERT represents SOTA architectures for computerized language analysis.

For this project, we fine-tune BERT-base to classify all four MBTI traits individually. We train the model separately for liked tweets, own tweets (statuses), and biographies. For liked tweets and statuses, all tweets (per user) are appended to one another and the first 512 tokens are sampled from this body to create the input text. For biographies, the first 64 tokens are sampled to create the input text (as biographies are much shorter than either the cumulative sum of liked tweets or statuses). The *BertForSequenceClassification* model from the Transformers Python package, published by Hugging Face, was used for all experiments. This model consists of a pre-trained BERT-base model with an additional classification layer with 2 nodes on top. The final models generated for each of the experiments differed in hyperparameter settings as the hyperparameters were tuned. The learning rate was set to $10^{-5}$, the value of epsilon to $10^{-8}$, the batch size varied between 1 to 8, and the number of epochs from 3 to 4. To create a performance baseline, a Naïve Bayes model was also run. The Naïve Bayes classification is based on token counts – input features are processed using a tf-idf vectorizer. The implementation of this model was obtained from the genism Python package.

### 3.3 Text embeddings and visualization

We perform visualizations of text embeddings of the tweets engaged by user type and the biographies of users. These embeddings are generated by taking the CLS token value (which is an aggregate representation of the input) from the outputs of the BERT model (before being fine-tuned). These embeddings are then visualized using UMAP. The motivation behind these visualizations is to provide insight into the topology of the text-space that the personality types engage in.

## 4. Results

### 4.1 BERT and Naïve Bayes Classifiers

| Trait | Acc. |
|---|---|
| Introverts / Extraverts | **65%** |
| Intuitive / Sensing | **73%** |
| Feeling / Thinking | **63%** |
| Prospecting / Judging | **63%** |

Table 2: Classification accuracy for the BERT trait classifiers using biographies, rounded to the nearest integer. All baselines 50%.

| Trait | Acc. |
|---|---|
| Introverts / Extraverts | **62%** |
| Intuitive / Sensing | **65%** |
| Feeling / Thinking | **68%** |
| Prospecting / Judging | **61%** |

Table 3: Classification accuracy for the Naïve Bayes trait classifiers for biographies, rounded to the nearest integer. All baselines 50%.

| Trait | Acc. |
|---|---|
| Introverts / Extraverts | **64%** |
| Intuitive / Sensing | **68%** |
| Feeling / Thinking | **63%** |
| Prospecting / Judging | **59%** |

Table 4: Classification accuracy for the BERT trait classifiers using statuses, rounded to the nearest integer. All baselines 50%.

| Trait | Acc. |
|---|---|
| Introverts / Extraverts | **55%** |
| Intuitive / Sensing | **63%** |
| Feeling / Thinking | **54%** |
| Prospecting / Judging | **51%** |

Table 5: Classification accuracy for the Naïve Bayes trait classifiers for statuses, rounded to the nearest integer. All baselines 50%.

| Trait | Acc. |
|---|---|
| Introverts / Extraverts | **60%** |
| Intuitive / Sensing | **67%** |
| Feeling / Thinking | **63%** |
| Prospecting / Judging | **59%** |

Table 6: Classification accuracy for the BERT trait classifiers using liked tweets, rounded to the nearest integer. All baselines 50%.

| Trait | Acc. |
|---|---|
| Introverts / Extraverts | **52%** |
| Intuitive / Sensing | **62%** |
| Feeling / Thinking | **57%** |
| Prospecting / Judging | **57%** |

Table 7: Classification accuracy for the Naïve Bayes trait classifiers for liked tweets, rounded to the nearest integer. All baselines 50%.

## 4.2 Text embedding visualizations using UMAP

### 4.2.1 Biographies

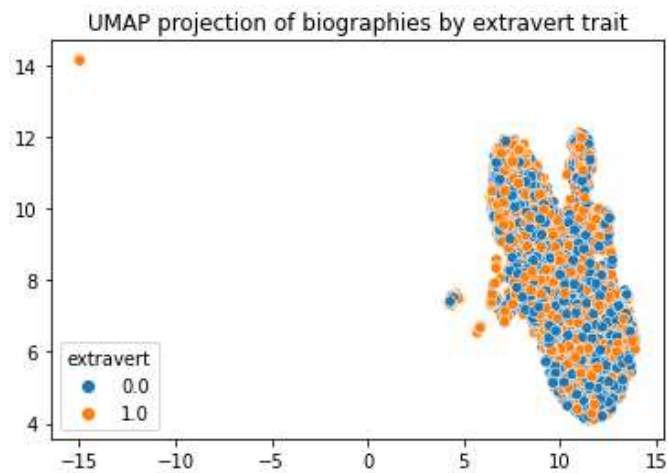

Fig 1.

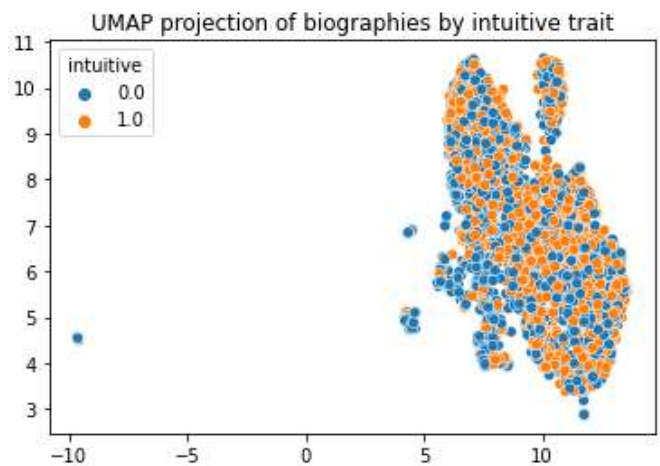

Fig. 2

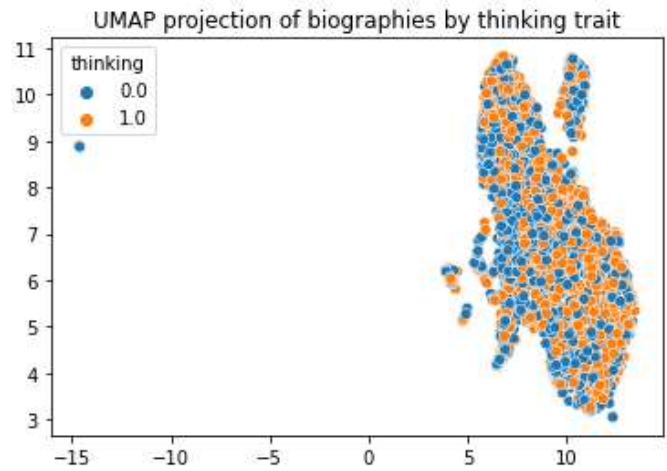

Fig. 3

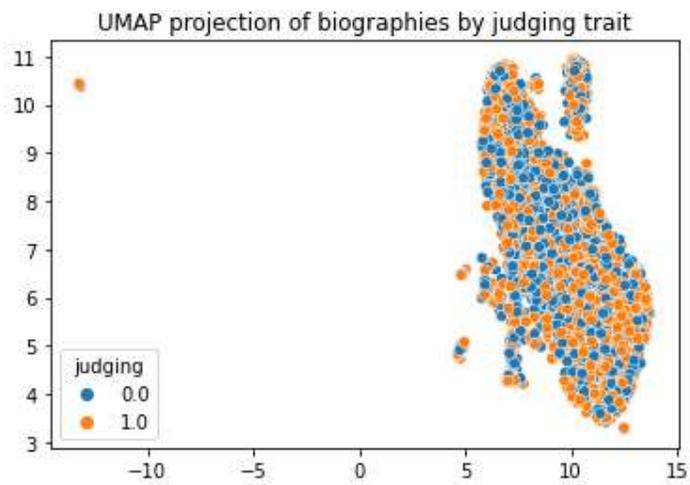

Fig. 4

### 4.2.2 Statuses

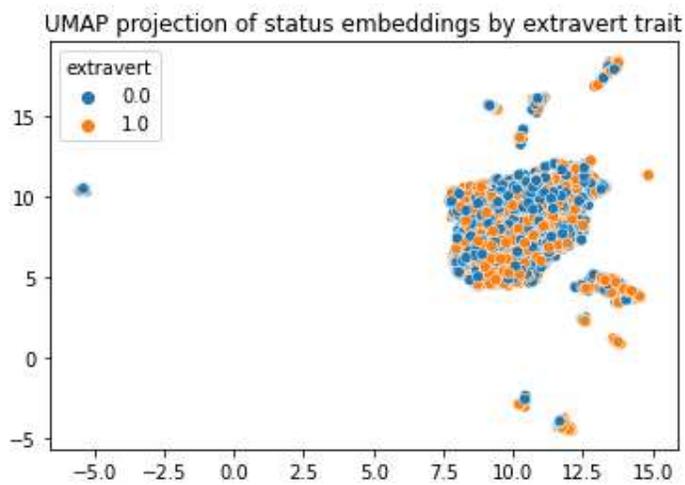

Fig. 5

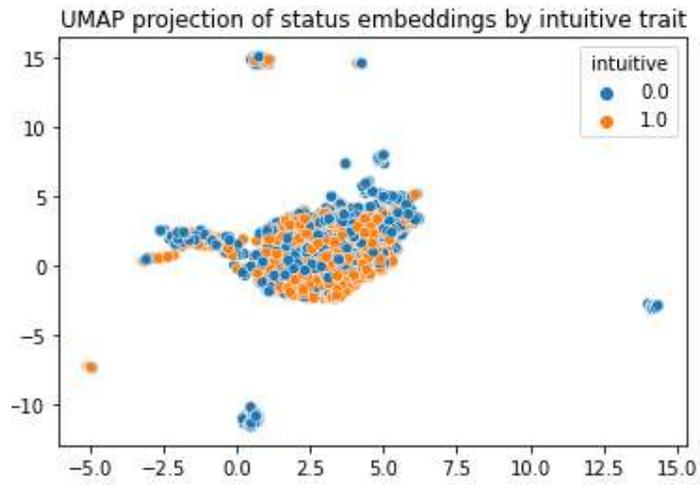

Fig. 6

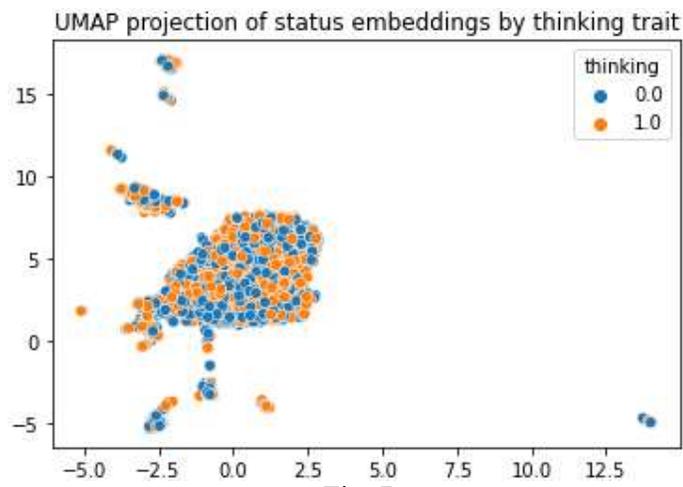

Fig. 7

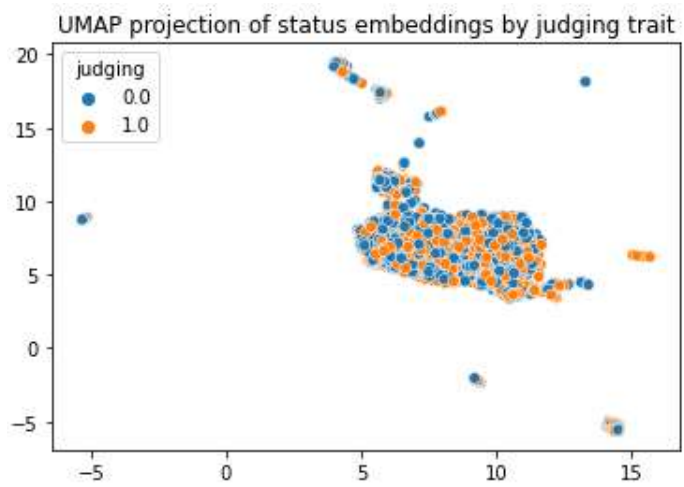

Fig. 8

### 4.2.3 Liked Tweets

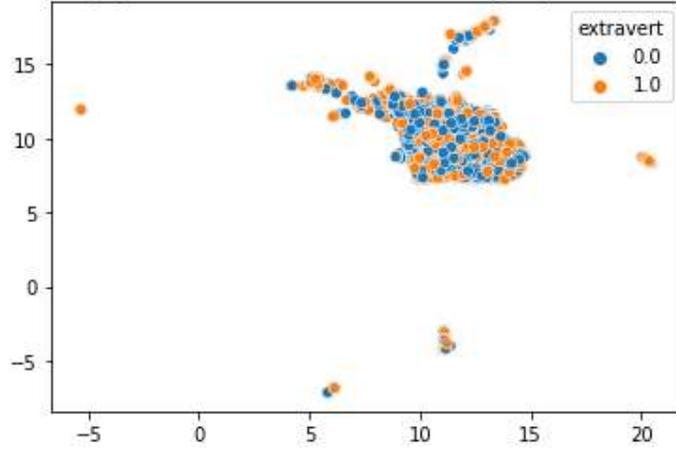

Fig. 9

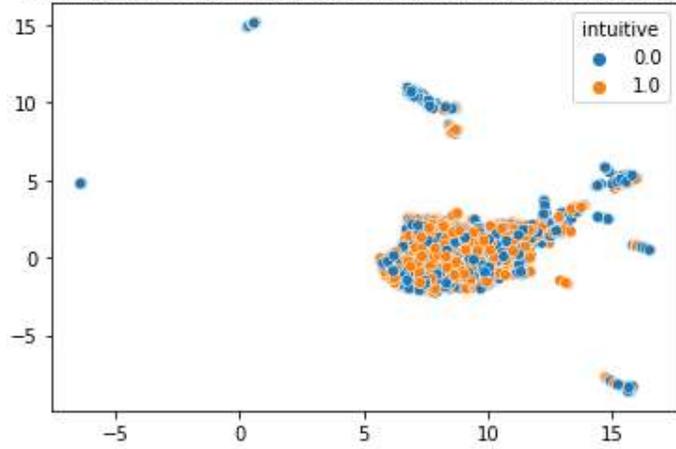

Fig. 10

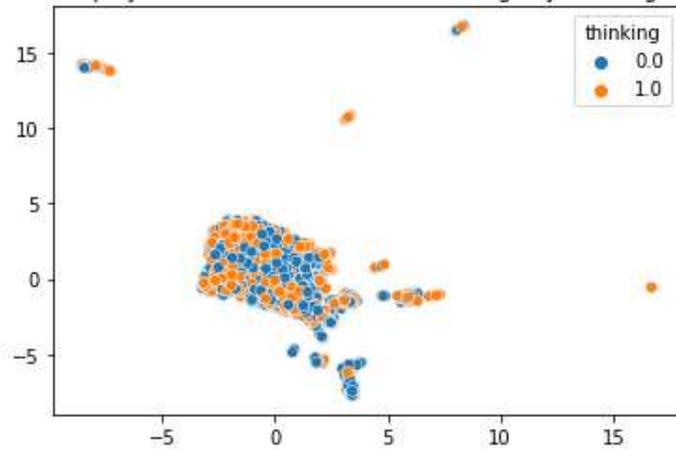

Fig. 11

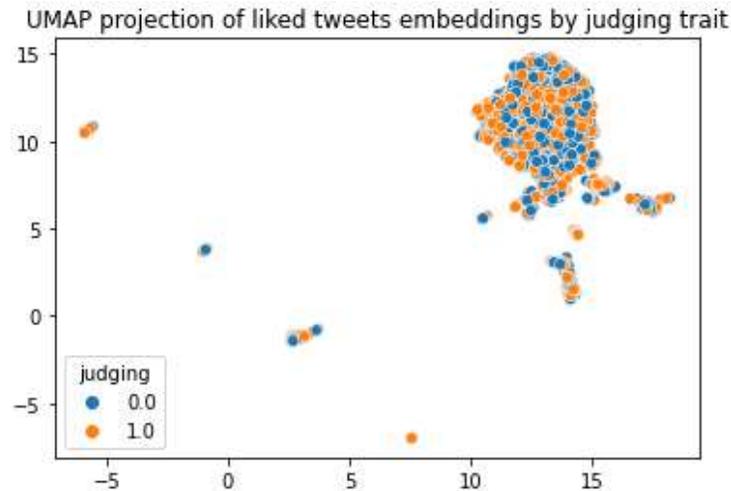
Fig. 12

## 5. Discussion

Both BERT and the Naïve Bayes – which serves as a performance baseline – perform better than chance on all experiments. BERT beats performance baselines on almost all model fits. We see that performance differs significantly across traits and features tested. While the median performance of all models for all tasks hovers around 60% - a modest figure – we cannot interpret the performance of these models on an absolute basis because it is not clear what the theoretically best possible accuracy for this task is – it is not 100%. The labels on the dataset are self-reported by a community of Myers-Briggs type enthusiasts, so it is unlikely that all of them are accurate in the sense that they would match labels generated through standardized testing. Even if users undertook professional tests to determine their types, survey-based measures for personality contain some degree of test-retest error. There is also biases that creeps in due to self-selection when choosing to self-report such labels in a public domain. In sum, we can be sure that the labels are not very accurate – even trained humans would be highly unlikely to perform close to perfect on these tasks.

We first analyze model performance on biographies – self-descriptions of the users. An intuitive prior would be that such text contains strong signal related to personality traits – and the results bear this out. Both BERT and Naïve Bayes perform significantly better than chance, with an accuracy up to 73% on the Intuitive/Sensing (N/S) dimension and nearly 2/3rds on other dimensions. Somewhat surprisingly, the far more simplistic Naïve Bayes outperforms BERT on the Thinking/Feeling dimension (68% vs 64%); the reason for this is unclear. On other traits, we see that Naïve Bayes performance is lower than that of BERT but is substantially better than chance. This indicates that a count-based approach that compares distributions of raw word counts captures significant information about personality but also fails to capture all relevant information. Note that biographies on Twitter often contain highly parsimonious descriptions of individuals i.e. 'Mother. Engineer. Soccer fan.' Biographies often don't contain fully-composed sentences and only singular word/phrasal descriptors. This could explain why Naïve Bayes only trails by just 2% on average compared to BERT on accuracy; BERT's significant advantage in *syntactic* parsing is not at play here.

Moving onto statuses, we see an especially large drop in Naïve Bayes performance – more than 8%, while BERT drops by an average of 2.5%. This drastic drop in Naïve Bayes accuracy is despite the increased availability of tokens from a maximum of 64 in biographies to 512 in statuses (which should lead to better estimates). This drop might be due to the feature change itself. Statuses, unlike biographies, are content generated by users that are not expected to be framed a certain way or address a particular theme. They are hence more indirect expressions of self. Statuses may contain more generic words which makes it more difficult to distinguish between two lexical distributions, whereas biographies contain tokens that are more unique – and consequently better distinguishing – on average. The superior syntactic parsing of BERT may be why it suffers less of a performance drop. In statuses too, we see that N/S stands out as the dimension with most predictive signal from text – it shows the highest accuracy across all traits for both methods.

Results from the liked tweets tasks paint a similar picture to that of statuses. However, liked tweets are still more indirect signals of self-expression than statuses – the users merely indicate that they find the content salient. These tweets can be interpreted as proxies for the information consumption preferences of users. Despite the somewhat incidental relationship between liked tweets and users, the predictive performance of the liked tweets features appears to be comparable to that of statuses. BERT performance drops only about 1% on average and Naïve Bayes performance increases by a few per cent. This may be because the average tokens available for liked tweets exceeds that of statuses (people like tweets far more often than they create tweets) – despite the numerical token cap of 512 applying to both features.

Broadly, these findings as a whole conform to intuitive priors we may have formed before we conducted the experiments. Descriptions of users, text content generated by users, and proxies for information consumption (liked tweets) all carry predictive signal corresponding to each of the personality dimensions we experiment on. Biographies – which are intended as descriptions of persons – carry most predictive power. Statuses – which are content generated by users – carry about as much signal as liked tweets, which are indirectly associated text. BERT, one of the most sophisticated model NLP models ever developed, unsurprisingly beats performance baselines consistently. Naïve Bayes baselines are comparatively high in the biographies task but lower in tasks containing features more indirectly related to user personality – tasks on which BERT suffers much less of a performance drop.

Through all experiments we see that the N/S dimension is the easiest to distinguish from text. Manual analysis may be able to easily reveal the features this dimension corresponds to. Note that the N/S dimension relates most strongly compared to the other three traits to the Five Factor Model of personality (where it is referred to as 'Openness'). Moreover, we see evidence from our raw dataset that 'N' users are overrepresented in our sample compared to a national average[23]; these findings and context strongly suggest that the N/S dimension is a significant axis of human behavioral diversity.

To obtain a better idea of how personality traits relate to the topology of the text-space occupied by users, we visualized CLS-token embeddings obtained from BERT using UMAP. CLS tokens are special tokens that capture an input-level aggregate representation using a 768 dimensional vector. BERT was not trained on the data before we obtained the CLS tokens for visualization. Our hope was that an 'unsupervised' visualization of user texts, labeled by personality traits, might reveal interesting patterns of homophily, echo-chambering, and other traces of sociological phenomena.

Unfortunately, our displays of the reduced two-dimensional representations of the CLS tokens colored by trait did not reveal any significant divisions, demarcation, or large-scale patterns between text and any of the personality traits for any of the features we tested on. This is not wholly unsurprising as just two components of a 768-dimensional dataspace might not capture enough variation in the data. While some small-scale features are visible in some of our graphs, there is not enough differentiation at large for us to be sure that they are not artefacts from our visualization or otherwise random occurrences. Note that some of the graphs differ between traits for the same text feature – this is because the data for the graphs are sampled again to ensure balanced classes for every trait.

## 6. Conclusion

Deep learning architectures for NLP like BERT have significant potential to help generate non-trivial insights into human behavior. This project demonstrates the feasibility of that proposition by showing that BERT can predict labels for behavior with a significant probability when trained on various text-based features mined from social media interaction.

While this project cannot draw strong conclusions regarding the validity of specific constructs of personality, it does demonstrate important findings. We demonstrate that deep-learning based NLP models – with their unprecedented ability to parse text - can empower observational studies of behavior. As the results bear out, our deep-learning model outperforms baseline models on almost all tasks. When used appropriately, they can also help validate social science theory and mine insights from text that contribute to theory building. This project demonstrates that *some* of the theoretical constructs underlying the Myers-Briggs personality system hold water to some degree. The performance of the Naïve Bayes model – which was not far behind BERT – also lends to the idea that the distributions of lexical items carry intrinsic psychobehavioral signals. We take the performance of these models, which rely solely on text based prediction, as further validation of recent studies[9,12] that argue that psychometric models based directly on text are principally sound approaches; these results support the lexical hypothesis in so far as psychological attitudes have *some* form of projection into everyday language.

Our evidence is strongest on the Intuitive/Sensing dimension – accuracy of up to 73%. This dimension is especially important as it correlates strongest (out of the four traits) with the Five Factor Model (to 'Openness'). The significance of other dimensions in our findings is by no means far behind – our best performing models, taken together, achieve an average accuracy of over 65%. While these numbers may still seem tepid, note that combining information across features and combining the models themselves will very likely lead to performance improvements.

Moreover, many large scale analyses or applications do not necessarily demand highly accurate classification to achieve their objectives. A study by Matz et al.[25] on personality-based targeting for advertising demonstrated that displaying tailor-made advertisements for persons using a dichotomous coding of the extraversion and openness dimensions – which is highly similar to the framework we use in this study – results in substantial increases in advertisement interest (up to 50% more purchases!). While they target populations based on Facebook likes, the purely text-based we develop could theoretical function in a similar capacity if validated further. As with text, the accuracy of likes-based identification of personality traits is far from perfect. For such applications and research, better-than-chance signal is often a sufficient criteria for real-world deployment.

There are many fronts on which this project can be expanded. While the predictive power of the experimental models we created improves on increasing the data available for training (though only slightly), larger improvements are realizable by taking a host of other data sources into consideration. These include profile pictures, shared images and GIFs, the tweets of the profiles that the users follow, and, in general, a more extensive accounting of the social network within which the user is situated. This project, using just biographies, statuses, and liked tweets, has likely only scratched the surface of what is possible.

While we discussed the possible sources of predictive power by comparing Naïve Bayes and BERT scores across multiple features, there is a possibility that Naïve Bayes and BERT are using mostly mutually exclusive feature sets to make predictions. If this is so, then an ensemble model composed of multiple classifiers and/or features would outperform BERT noticeably. This possibility has not been investigated in this project but strongly warrants further investigation.

Finally, it would be interesting if the same approach would be applied to data from a different social media platform – this would serve to validate the findings and the approach adopted in this project.